\documentclass[conference]{IEEEtran}
\IEEEoverridecommandlockouts

\usepackage{graphicx} 
\usepackage{amsmath,amssymb,amsfonts}
\usepackage[algo2e]{algorithm2e}
\usepackage{algpseudocode}
\usepackage{algorithm}
\usepackage{graphicx}
\usepackage{lipsum}
\usepackage{siunitx}
\usepackage{cite}
\usepackage[left=0.625in,right=0.625in,bottom=0.90in,top=0.9in]{geometry}

\title{Learning Power Control Protocol for In-Factory 6G Subnetworks}
\author{
\IEEEauthorblockN{Uyoata E. Uyoata, Gilberto Berardinelli, Ramoni O. Adeogun}
\IEEEauthorblockA{Wireless Communication Networks Section, Department of Electronic Systems, Aalborg University, Denmark\\ Email: \{ueu,gb, ra\}@es.aau.dk}
\thanks{This work is supported by the HORIZON-JU-SNS-2022-STREAM-B-01-02 CENTRIC project (grant agreement No. 101096379). The work by Ramoni Adeogun and Gilberto Berardinelli is also supported by the HORIZON-JU-SNS-2022-STREAM-B-01-03 6G-SHINE project (grant agreement No. 101095738). }}

\begin{document}

\maketitle
\begin{abstract}
In-X Subnetworks are envisioned to meet the stringent demands of short-range communication in diverse 6G use cases. In the context of In-Factory scenarios, effective power control is critical to mitigating the impact of interference resulting from potentially high subnetwork density. Existing approaches to power control in this domain have predominantly emphasized the data plane, often overlooking the impact of signaling overhead. Furthermore, prior work has typically adopted a network-centric perspective, relying on the assumption of complete and up-to-date channel state information (CSI) being readily available at the central controller. This paper introduces a novel multi-agent reinforcement learning (MARL) framework designed to enable access points to autonomously learn both signaling and power control protocols in an In-Factory Subnetwork environment. By formulating the problem as a partially observable Markov decision process (POMDP) and leveraging multi-agent proximal policy optimization (MAPPO), the proposed approach achieves significant advantages. The simulation results demonstrate that the learning-based method reduces signaling overhead by a factor of 8 while maintaining a buffer flush rate that lags the ideal "Genie" approach by only 5$\%$.


\end{abstract}

\section{Introduction}
The development of sixth-generation (6G) wireless communication technologies is a key focus for network operators, standardization bodies, and researchers. To support low-power, short-range communication at the network edge, in-X subnetworks have been proposed \cite{9585402, adeogun2020towards}. These subnetworks are designed for diverse use cases such as In-body, In-Factory, and In-vehicle scenarios and are typically densely deployed on entities of interest, such as robot arms or within the human body. This dense deployment results in interference, particularly when radio resources are shared among subnetworks, necessitating novel resource management strategies tailored to such subnetworks \cite{10118984, adeogun2024federated,10571230, adeogun2023unsupervised}.

For In-Factory scenarios, power control is critical to mitigate interference due to high subnetwork density. A graph neural network-based approach proposed in \cite{10118984} centrally manages transmit power by using channel state information (CSI) and inter-subnetwork distances to represent the network as a graph. This method improves spectral efficiency by approximately $5\%$ over a weighted minimum mean squared error (WMMSE)-based power control benchmark. However, it assumes full CSI availability at the central controller and requires subnetworks to send CSI updates at every time step, leading to high signaling overhead. Additionally, it does not account for the dynamic nature of factory modules, assuming instead that nodes remain stationary. Sequential Iterative Power Allocation (SIPA) and Gradient Descent Power Allocation (GDPA) methods proposed in \cite{10571230} were shown to improve network performance over fixed power strategies. However, both approaches also assume frequent CSI updates for every transmission cycle and rely on predefined signaling protocols, which may not always be optimal or practical.

The integration of machine learning into wireless communication has shown promise for protocol design and optimization \cite{10624788, 10000805, 10288545}. For instance, the work in  \cite{article} formulates a decentralized partially observable Markov decision process (Dec-POMDP) to enable user equipment (UE) agents to learn channel access and signaling policies. Reported results demonstrate that over extended training episodes, UE agents can adapt to standard signaling policies while optimizing rewards. Similarly, the work in  \cite{10437954} uses a multi-agent proximal policy optimization (MAPPO) framework where industrial IoT devices learn channel access, signaling, and task offloading decisions to maximize offloaded tasks within a delay budget. Simulation results confirm that MAPPO outperforms benchmark strategies. In another example, the authors in \cite{10624788} propose a proximal policy optimization (PPO) framework to optimize MAC protocol components of IEEE 802.11ac, achieving improved throughput and delay performance by dynamically adjusting protocol operations.

Although the foregoing research efforts demonstrate the potential of learning-based approaches, their focus has not been on In-Factory subnetworks, moreover, some have  assumed the existence of predefined signaling protocols. Such an assumption may not completely describe the envisaged dynamic In-Factory environments. In such a context, signaling protocols may need to adapt to dynamic environments without always relying on human-crafted solutions. Motivated by this, our proposed approach focuses on evolving protocols tailored to In-Factory subnetworks. Specifically, our goal is to maximize the number of downlink packets successfully transmitted by access points (APs) while meeting a link rate threshold. Our approach does not assume predefined signaling protocols but instead enables agents to learn policies through interaction with the dynamic wireless environment, significantly reducing signaling overhead.

\section{System Model }

\begin{figure} [t]
    \centering
    \includegraphics[width=1.1\linewidth]{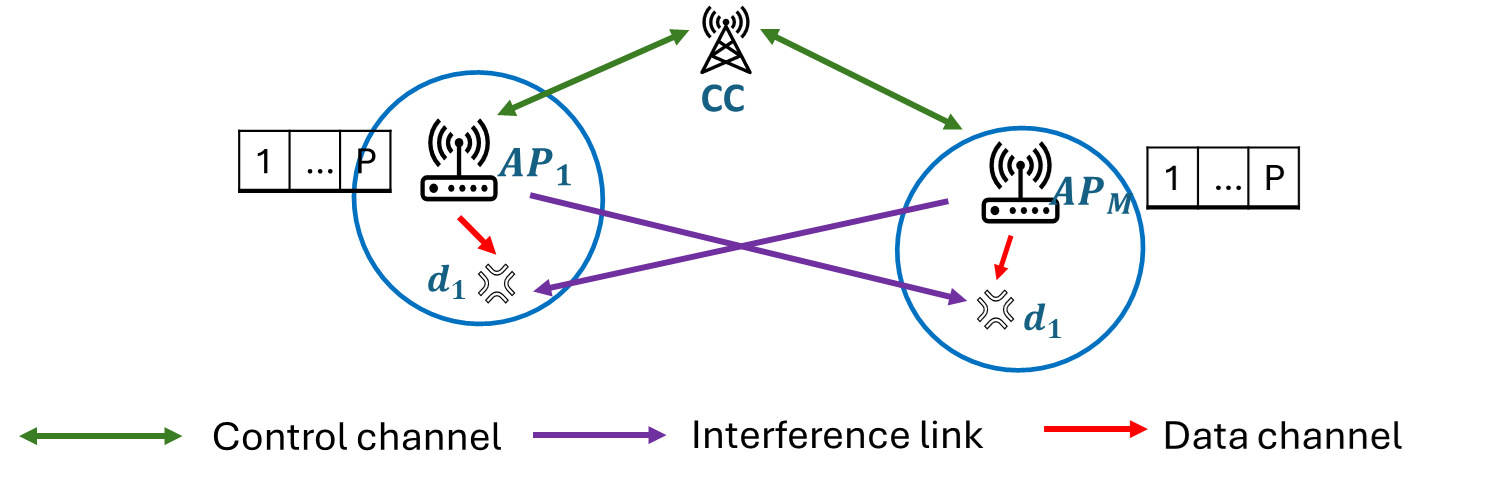}
    \caption{A network of M subnetworks showing a CC,  access points (APs) and  sensor/actuator unit. Signaling link are indicated by green arrows, measurements are indicated by the the blue arrow, control signals are indicated by the red arrows. Each AP has a maximum of P packets in its buffer.}
    \label{f1}
\end{figure}

We consider a network of $M$ subnetworks indexed by  $ m \in \mathcal{M} = \{1,2..., M \}$. Each subnetwork has an access point (co-located with a controller) that coordinates  $N$ sensor/actuator units indexed by $ n \in \mathcal{N} = \{1,2..., N \}$ . 
A central controller/base station provides channel access for the subnetworks as shown in Figure~\ref{f1}. APs exchange signaling messages with the central controller (CC) through error-free control channels, and send downlink command carrying packets to their associated devices through a data channel. Time Division Multi Access (TDMA) is assumed such that the total transmission duration is divided into slots. It is assumed that the central controller assigns orthogonal frequency resources for transmission in each subnetwork and so each subnetwork is inter-subnetwork interference limited. 

Our target is to train RL agents to learn the signaling protocol for an in-Factory subnetwork scenario in a way that meets a communication link rate constraint.



\section{ RL based learned protocol for power control}
The power allocation protocol learning problem is formulated as a multi-agent reinforcement learning (MARL) task in which APs learn to: use their signaling message, interpret the signaling messages from the CC and send command carrying packets to their associated devices. To model the formulated problem, a communication enhanced decentralised partially observable Markov decision process (POMDP) is employed as in \cite{Mota2021TheEO, 10437954}. For example, for $M$ AP agents, the POMDP is characterised by a global state space ($\mathcal{S}$), an action space ($\mathcal{A}=\mathcal{A}_{1},\mathcal{A}_{2},...,\mathcal{A}_{M}$) and an observation space ($\mathcal{O} = \mathcal{O}_{1},\mathcal{O}_{2},...,\mathcal{O}_{M}$). We consider the CC to be an expert agent that does not need to learn a power control policy. Sensors/Actuators are not learning agents as well.

\subsection{Agent Observation Space}
Both the $M$ AP agents and the CC have their observations from the shared environment. Each learning cycle is divided into equal-sized time steps, $t$. In each time step $t$, the observation for the $m_{th}$ AP agent is given by the number of queued command carrying packets in its buffer, i.e, $\mathrm{o}^{m}_{t} = |\mathcal{C}|$, where $\mathcal{C} = \{ 1,2,...,C\}$ and $C$ is the maximum number of packets an AP buffer can take.

In the same vein, the observation for the CC is given by the channel state information (CSI) reports of all $M$ subnetworks, i.e. $\mathrm{o}^{cc}_{t} = \begin{bmatrix}
h_{1,1} & . & .&h_{1,m}\\
. & . &. &.\\
. & .& .&.\\
h_{m,1} & . & .&h_{m,m}
\end{bmatrix}$, where the diagonal entries are the desired channel gains, i.e. the channel gains between an AP and the devices in its subnetwork whereas the other entries are channel gains of interfering signals. The considered model assumes a single device in a subnetwork hence the dimension of CSI matrix. The signal to interference plus noise ratio (SINR) for the desired link between a device and the AP in $m_{th}$ subnetwork is given by:

\begin{equation}
     \gamma_{m,m} = \dfrac{p_{m}h_{m,m}}{\sum\limits_{\substack{l=1, \\ l\neq m}}^{M} p_{l}h_{l,m}+ \sigma^{2}}.
     \label{eq2}
\end{equation}
In equation \ref{eq2}, $p_{m}$ is the transmit power of the $m_{th}$ AP, $h_{m,m}$ is the desired signal within the $m_{th}$ subnetwork and it comprises the large scale path loss, small scale fading and shadowing. $h_{l,m}$ is the channel gain of the interference from the $l_{th}$ subnetwork and $p_{l}$ is the transmit power of the AP in the $l_{th}$ subnetwork that shares the same frequency resource with the $m_{th}$ subnetwork. The thermal noise power at each receiving sensor/actuator within the $m_{th}$ subnetwork is given by  $\sigma^{2} = KTB\times 10^{NF/10}$ where $K$ is the Boltzmann constant, $T$ is the noise temperature in Kelvin, $NF$ is the noise figure  and $B$ is the system bandwidth in \si{MHz}. $I_{m} =\sum\limits_{\substack{l=1, \\ l\neq m}}^{M} p_{l}h_{l,m}$ is the sum interference experienced by the device in the $m_{th}$ subnetwork.
Therefore the link rate of the desired link in the  $m_{th}$ subnetwork is given as:
\begin{equation}
   R_{m,m} \approx B\log(1 + \gamma_{m,m}).
\end{equation}
For transmitted packet in a subnetwork to be considered successful, $R_{m,m}$ must exceed a threshold rate $R_{th}$ and its experienced delay should be less than a delay deadline, $\tau_{th}$.

The delay experienced by a packet transmitted by the $m_{th}$ AP is then given by,

\begin{equation}
   t^{m} = \dfrac{A}{R_{m,m}} + \tau^{m} 
\end{equation}
where $A$ is the packet size, $\tau^{m} $ is the length of time in seconds spent by the packet in the AP buffer. Hence, a packet transmission is successful if $R_{m,m} \geq R_{th}$.

\subsection{Agent Action Space}
Each agent has two action spaces depending on the effect of a selected action on the environment: the communication action does not have direct impact on the environment whereas the environment action directly affects the environment.

In each time step $t$, the environment action which the $m_{th}$ AP can take is given by $a^{m}_{t} \in \{0,1\}$  where $0$ means AP does nothing, $1$ means AP transmits command carrying packet.
Similarly, the communication actions of the $m_{th}$ AP are control messages sent to the CC in the uplink and is given by $U^{m}_{t} = \{0,1\}$ where $0$ means AP sends Power Allocation Request (PAR), $1$ means AP sends CSI report. Due to the overhead incurred by frequent CSI reports, the AP agents will need to learn when using this communication action is pertinent. 

Similarly, the CC also sends communication messages to each $m$ AP given by $D^{m}_{t} = \{0,...,1\}$ where  $\{0,..,1\}$ indicates the coefficient of power allocation. The CC being an expert agent understands the meaning of both uplink and downlink communication messages. However, each AP agent will need to learn how to use the uplink communication messages based on its interaction with the environment. 

\subsection{Agent State Space}
 At time step $t$, the state ($x^{m}_{t}$) of the $m_{th}$ agent consists of the $q$ observations, actions and its received communication messages up to that time. That is,
$x^{m}_{t} = (o^{m}_{t},..., o^{m}_{t-q}, a^{m}_{t},..., a^{m}_{t-q}, , U^{m}_{t},..., U^{m}_{t-q}, D^{m}_{t},..., D^{m}_{t-q}, )$.
For the expert agent (i.e. the CC), the state at time $t$ is given by, $x^{CC}_{t} = (o^{CC}_{t},..., o^{CC}_{t-q} , \textbf{U}_{t},..., \textbf{U}_{t-q}, \textbf{D}_{t},..., \textbf{D}_{t-q}, )$where \textbf{U} and \textbf{D} contains uplink and downlink messages respectively for all APs. The state of the environment, $s_{t} \in \mathcal{S}$ combines both $x^{m}_{t}$ and $x^{CC}_{t}$.
\subsection{Power Allocation Algorithm}
At the beginning of communication, the CC allocates a fixed transmit power to all subnetworks. In subsequent time steps, APs can choose to send power allocation requests.
The power allocation algorithm used in the CC or BS is based on graph neural network (GNN) details of which are given in \cite{10118984}.



\subsection{Agent Reward}
At each time step, each AP agent receives a reward ($r_{t}^{m}$) based on its interaction with the environment. This reward is an average of the total reward for all agents, i,e. $r_{t}^{m} =\dfrac{1}{M} \sum_{m=1}^{M}  r_{t}^{m}$ where,  

\begin{equation}
    r_{t}^{m}= \begin{cases} +\mathrm{z} , & \text{if tx succeeds, (i.e., $R_{m,m} \geq R_{th}$)} \\
    0 & \text{tx fails }\\
0 &  \text{otherwise} \\
    \end{cases}
\end{equation}. 
The agent is awarded a positive value ($+z$) when there is a successful transmission to encourage agents toward transmission. However, failed transmissions are punished. 

\section{Performance Evaluation}
We consider subnetworks located on a $10m \times 10m$ area. Each AP, device and the  CC are equipped with a single antenna. We use similar indoor factory scenario as in \cite{10817320} where each subnetwork moves at a speed of \SI{3}{m/s} in defined paths without collision. Each subnetwork has a radius of \SI{0.5}{m}. Network configurations for simulation are given in Table \ref{T2} and the parameters for the MAPPO configuration are given in Table \ref{T3}. 
\begin{table}
\caption{Network Configuration Parameters} 
\centering 
\begin{tabular}{l|l} 
\hline 
\textbf{Parameters} & \textbf{Values} \\
\hline
Deployment area & 10m $\times$ 10m  \\
\hline
Subnetwork radius & 1m\\
\hline
Operating frequency& \SI{6}{GHz} \\
\hline
Bandwidth& \SI{10}{MHz} \\
\hline
Path loss exponent & $2.7$\\
\hline
Maximum speed of subnetworks & \SI{3}{m/s}\\
\hline
Maximum transmit power & \SI{20}{dBm} \\
\hline
Minimum transmit power & \SI{0}{dBm}\\
\hline
Number of subnetworks & $10$\\
\hline
Number of devices per subnetwork & $1$\\
\hline
Distance between device and and AP & \SI{0.5}{m}\\
\hline
Noise spectral density & \SI{-174} {dBm/Hz}\\
\hline
Payload & $64$ bytes\\
\hline
Latency & $0.001$s\\
\hline
Threshold Spectral Efficiency ($R_{th}$)& \SI{0.05}{bps/Hz}\\
\hline
Capacity of AP buffer& 100\\
\hline
\end{tabular}
\label{T2}\vspace{-10pt}
\end{table}

\begin{table}
\caption{RL Parameter configuration} 
\centering 
\begin{tabular}{|l|l|l|l|} 
\hline 
\textbf{Parameters} & \textbf{Values} & \textbf{Parameters} & \textbf{Values}\\
\hline
Learning rate &$0.0001$  &  Discount factor & $0.99$\\
\hline
PPO epochs& 2 & Number of episodes& $1000$\\
\hline
Optimizer& Adam & Steps per episode&$300$ \\
\hline
Max. memory length&$64000$  & Sample frequency&$0.05$\\
\hline
Clipping parameter& $0.2$ & Mini-batch size&$256$\\
\hline
\end{tabular}
\label{T3}\vspace{-10pt}
\end{table}

\subsection{MAPPO Architecture}
To solve the above formulated problem, we use multi-agent proximal policy optimization (MAPPO) algorithm. Proximal policy optimization (PPO) algorithm achieves better training stability by gently updating the policy per training episode, and so guarantees training convergence. PPO optimizes a surrogate objective function ($\mathcal{L(\theta)}$) which is a function of the ratio between the current policy and the old policy and clips this ratio within a specified range.

\begin{equation}
    \mathcal{L}_{t}(\theta) = \mathbb{E}[min(\zeta_{t}(\theta),clip(\zeta_{t}(\theta), 1-\epsilon, 1 + \epsilon)\mathrm{A}_{t})],
    \label{eq_surr}
\end{equation}
where $\zeta(\theta)$ denotes the probability of taking an action using the current policy ($\pi_{\theta}$) and the old policy ($\pi_{\theta_{old}}$), $\epsilon$ determines the limit of clip variation and $\mathrm{A}_{t}$ is the advantage function which scores the state value based on the reward returned for an action.
PPO consists of the value/critic network and the action/policy network. The actor network outputs the probability distribution over the possible actions given the state of the environment. Given a state and the action of the agent, the critic network outputs the value of the cumulative reward of the given state, action pair.

For the work in this paper, both the actor and critic networks use a sequential model having four linear layers and \textit{Tanh} activation functions. To achieve an output of probability distributions, a softmax activation function is added to the output layer of the actor network. The logical flow of our proposed learning power control protocol is given in Algorithm 1.
\begin{algorithm}
\caption{: MAPPO based learning power control protocol}
\textbf{Initialize:} learning rate $\eta$, discount factor $\gamma$, mini batch size $M$, actor network  $\theta_{a}$, critic network $\theta_{c}$, experience buffer $Bf$ \\
\For {episodes, v = 1, 2,..., V}
{
Intialize subnetwork position and speed to get initial state\\
\For {time step, t = 1, 2,..., T}
{
Obtain the AP observations, $o^{m}_{t}$ and CC observation, $o^{CC}_{t}$\\
Select action, $a_{t}$  based on policy, $\pi_{\theta_{a}}$ \\
Apply action, $a_{t}$, observe reward, $r_{t}$ and get new state, $s_{t+1}$\\
Save tuple $\langle s_{t}, a_{t},   r_{t}, a_{t+1} \rangle$ in the buffer $Bf$ of each agent
}
Compute Advantage functions, $\mathrm{A}_{t}$\\
Compute policy probability ratios, $\zeta_{t}$ \\
Compute surrogate objective function based on equation (\ref{eq_surr})  \\
Compute value function loss
}
\Return Optimal power control policy 
\end{algorithm}
\subsection{Benchmarks} 
Our proposed technique will be compared to the following benchmark approaches:
\subsubsection{Fixed Action} In this scheme, APs transmit in every time step with a constant power (\SI{20}{dBm}) irrespective of the packet reception status. CSIs and power allocation requests are not sent by the APs.
\subsubsection{Random Action} AP agents choose to either do nothing, transmit packets, send CSI or power allocation requests. Basically, APs select an action in a random manner without recourse to feedback from the environment. When an agent selects to transmit a packet, it transmits using a transmit power allocated by the central controller in the last power allocation round. When an AP selects to send a power allocation request, the CC performs power allocation using a GNN based power allocation strategy.

\subsubsection{Interference-Aware} In this scheme, AP agents  initially transmit at a fixed power, $p_{0}$ as long as transmitted packets are received at or above the threshold rate ($R_{th}$). When a packet is not received correctly, APs send a CSI update to the CC, followed by a power allocation request. Subsequent AP transmissions use newly allocated power levels. Power allocation request and CSI updates are not sent concurrently, hence two transmission opportunities are missed.

\subsubsection{CSI Based Interference-Aware } This approach is similar to the Interference-Aware approach but differs in the frequency of CSI report transmission. Whereas in the Interference-Aware approach, AP agents only send CSI updates if their transmission fails, in the CSI based Interference-Aware Approach, CSI updates are sent in every time step to the CC. The availability of updated CSI  guides the CC in making concise power allocation decision.
\subsubsection{Persistent Power} This is a 'Genie' approach wherein the APs send CSI updates in every time step to the CC and in every time step, power allocation is performed. No transmission opportunity is missed in this approach.
\begin{figure}[htb]
    \centering
    \includegraphics[width=0.85\linewidth]{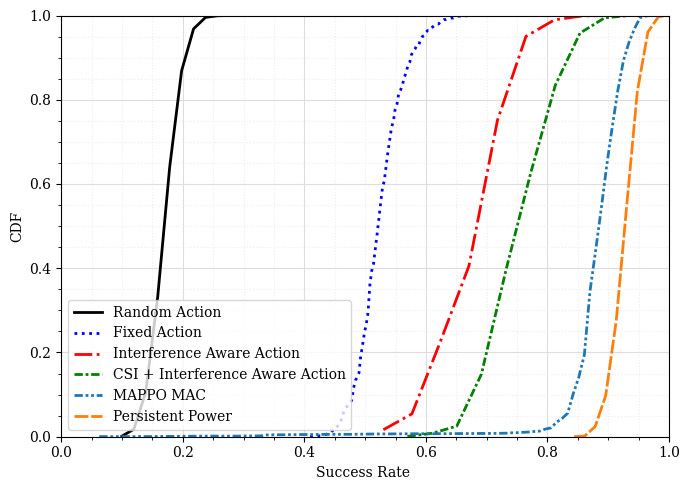}
    \caption{CDF of success rate of the considered approaches}
    \label{f3}\vspace{-10pt}
\end{figure}
\begin{figure}[!h]
    \centering
\includegraphics[width=0.75\linewidth]{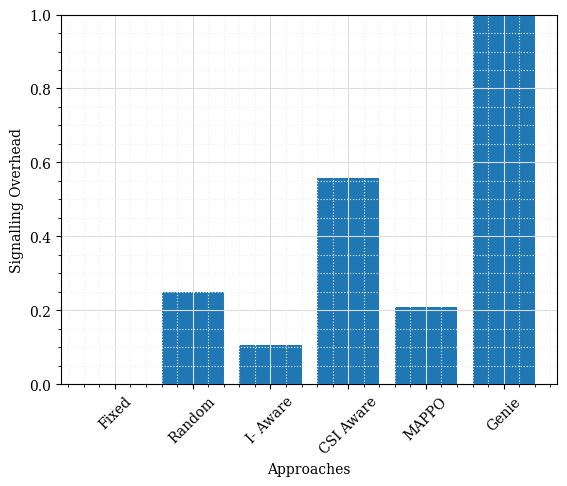}
    \caption{Signaling overhead of the considered approaches}
    \label{f4}\vspace{-10pt}
\end{figure}
\begin{figure} [t]
    \centering
\includegraphics[width=0.85\linewidth]{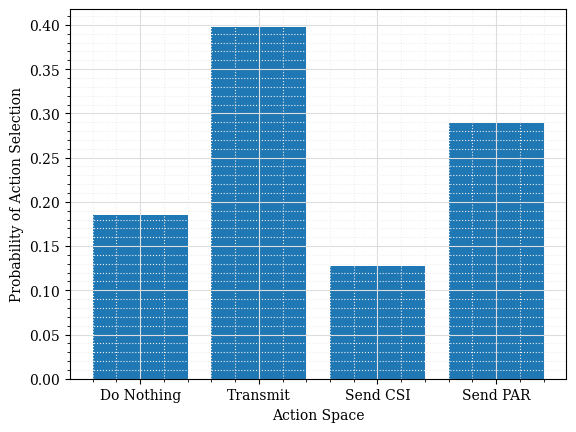}
    \caption{Action Selection Probability}
    \label{f5}\vspace{-10pt}
\end{figure}
\subsection{Results}
In this section, we present the results of simulating both the proposed learning-based approach and the benchmark approaches. The benchmarks were executed on a Lenovo X1 Carbon laptop with an Intel Core i7 processor and 32GB of RAM, while the learning-based simulations utilized remote computing clusters to accommodate the computational demands of reinforcement learning. $1000$ simulation episodes were carried out with each episode consisting of $300$ steps.

\subsubsection{Success Rate}
Figure~\ref{f3} illustrates the  cumulative distribution function (CDF) of the success rates of the proposed approach and the benchmark methods. For this study, the success rate is defined as,
\[
\text{Success Rate} = \frac{N_{\text{transmissions}}}{\min(T_{\text{flush}}, N_{\text{steps}})},
\]
where $N_{\text{transmissions}}$ represents the total number of transmitted packets, $T_{\text{flush}}$ is the time at which an access point (AP) empties its buffer, and $N_{\text{steps}}$ denotes the total number of steps in an episode. The figure shows that the Genie approach achieves a success probability of approximately $0.9$ at the $50^{th}$ percentile, indicating that $50\%$ of the time, around $90\%$ of the buffered packets are successfully transmitted. In contrast, the \textit{Random Action} approach has a success rate of about $0.18$ at the $50^{th}$percentile, which aligns with expectations since agents in this method choose to transmit only one out of four times, losing $75\%$ of transmission opportunities. The proposed MAPPO-based approach achieves a success rate of approximately $0.85$ at the $50^{th}$ percentile, only $0.05$ below the Genie approach. This demonstrates that the learning-based approach closely matches the performance of the optimal method while relying solely on learned policies.

\subsubsection{Signaling Overhead}
Figure~\ref{f4} examines the signaling overhead of the evaluated approaches. The \textit{Persistent Power} approach incurs the highest signaling overhead, as expected. The \textit{CSI-based Interference Aware} approach also exhibits significant overhead, approximately $8\times$ higher than the \textit{Interference Aware} method, further emphasizing the cost associated with frequent CSI updates. Note that in obtaining the signaling overhead, we summed each incidence of signaling over the entire simulation time.

Although the proposed learning-based approach does not achieve the lowest signaling overhead, it reduces signaling costs by approximately $8\times$ compared to the Genie approach. This substantial reduction highlights the effectiveness of the MAPPO framework in minimizing the frequency of signaling interactions while maintaining high performance.

\subsubsection{Action Selection Probability}
Figure~\ref{f5} depicts the action selection probabilities of the MAPPO agents. The results reveal that power allocation requests contribute significantly to the signaling overhead of the MAPPO-based approach. Although the framework did not explicitly penalize frequent CSI updates during training, the agents were able to adapt and learn to send fewer CSI updates autonomously. This behavior demonstrates the ability of the proposed approach to optimize signaling patterns in response to environmental dynamics and task requirements.

Overall, the proposed MAPPO-based approach strikes a balance between achieving high success rates and reducing signaling overhead, making it a practical solution for In-Factory subnetwork scenarios.

\section{Conclusion}
This paper presented a novel formulation of the power allocation protocol learning problem, specifically tailored to In-Factory subnetworks, and proposed a Multi-Agent Proximal Policy Optimization (MAPPO) framework as a solution. By leveraging reinforcement learning, the proposed approach enables agents to autonomously learn efficient signaling and power allocation strategies in dynamic and resource-constrained environments. Simulation results demonstrate that the MAPPO-based approach achieves a significant reduction in signaling overhead compared to conventional benchmarks, while maintaining a high success probability that closely approaches the optimal "Genie" solution. 

\bibliographystyle{IEEEtran}
\bibliography{mac_ref}

\begin{thebibliography}{10}
\providecommand{\url}[1]{#1}
\csname url@samestyle\endcsname
\providecommand{\newblock}{\relax}
\providecommand{\bibinfo}[2]{#2}
\providecommand{\BIBentrySTDinterwordspacing}{\spaceskip=0pt\relax}
\providecommand{\BIBentryALTinterwordstretchfactor}{4}
\providecommand{\BIBentryALTinterwordspacing}{\spaceskip=\fontdimen2\font plus
\BIBentryALTinterwordstretchfactor\fontdimen3\font minus \fontdimen4\font\relax}
\providecommand{\BIBforeignlanguage}[2]{{%
\expandafter\ifx\csname l@#1\endcsname\relax
\typeout{** WARNING: IEEEtran.bst: No hyphenation pattern has been}%
\typeout{** loaded for the language `#1'. Using the pattern for}%
\typeout{** the default language instead.}%
\else
\language=\csname l@#1\endcsname
\fi
#2}}
\providecommand{\BIBdecl}{\relax}
\BIBdecl

\bibitem{9585402}
G.~Berardinelli, P.~Baracca, R.~O. Adeogun, S.~R. Khosravirad, F.~Schaich, K.~Upadhya, D.~Li, T.~Tao, H.~Viswanathan, and P.~Mogensen, ``Extreme communication in {6G}: Vision and challenges for ‘in-{X}’ subnetworks,'' \emph{IEEE Open Journal of the Communications Society}, vol.~2, pp. 2516--2535, 2021.

\bibitem{adeogun2020towards}
R.~Adeogun, G.~Berardinelli, P.~E. Mogensen, I.~Rodriguez, and M.~Razzaghpour, ``Towards {6G} in-{X} subnetworks with sub-millisecond communication cycles and extreme reliability,'' \emph{IEEE Access}, vol.~8, pp. 110\,172--110\,188, 2020.

\bibitem{10118984}
D.~Abode, R.~Adeogun, and G.~Berardinelli, ``Power control for {6G} industrial wireless subnetworks: A graph neural network approach,'' in \emph{2023 IEEE Wireless Communications and Networking Conference (WCNC)}, 2023, pp. 1--6.

\bibitem{adeogun2024federated}
R.~Adeogun, ``Federated deep deterministic policy gradient for power control in {6G} in-{X} subnetworks,'' in \emph{2024 14th International Conference on Electrical Engineering (ICEENG)}.\hskip 1em plus 0.5em minus 0.4em\relax IEEE, 2024, pp. 199--201.

\bibitem{10571230}
D.~Li, S.~R. Khosravirad, T.~Tao, P.~Baracca, and P.~Wen, ``Power allocation for {6G} sub-networks in industrial wireless control,'' in \emph{2024 IEEE Wireless Communications and Networking Conference (WCNC)}, 2024, pp. 1--6.

\bibitem{adeogun2023unsupervised}
R.~Adeogun, ``Unsupervised deep unfolded {PGD} for transmit power allocation in wireless systems,'' in \emph{2023 IEEE 34th Annual International Symposium on Personal, Indoor and Mobile Radio Communications (PIMRC)}.\hskip 1em plus 0.5em minus 0.4em\relax IEEE, 2023, pp. 1--5.

\bibitem{10624788}
N.~Keshtiarast and M.~Petrova, ``{ML} framework for wireless {MAC} protocol design,'' in \emph{2024 IEEE International Conference on Machine Learning for Communication and Networking (ICMLCN)}, 2024, pp. 560--565.

\bibitem{10000805}
L.~Miuccio, S.~Riolo, S.~Samarakoon, D.~Panno, and M.~Bennis, ``Learning generalized wireless {MAC} communication protocols via abstraction,'' in \emph{GLOBECOM 2022 - 2022 IEEE Global Communications Conference}, 2022, pp. 2322--2327.

\bibitem{10288545}
J.~Xiao, Z.~Chen, X.~Sun, W.~Zhan, X.~Wang, and X.~Chen, ``Online multi-agent reinforcement learning for multiple access in wireless networks,'' \emph{IEEE Communications Letters}, vol.~27, no.~12, pp. 3250--3254, 2023.

\bibitem{article}
A.~Valcarce and J.~Hoydis, ``Toward joint learning of optimal {MAC} signaling and wireless channel access,'' \emph{IEEE Transactions on Cognitive Communications and Networking}, vol.~PP, pp. 1--1, 05 2021.

\bibitem{10437954}
S.~Mostafa, M.~P. Mota, A.~Valcarce, and M.~Bennis, ``Emergent communication protocol learning for task offloading in {I}ndustrial {I}nternet of {T}hings,'' in \emph{GLOBECOM 2023 - 2023 IEEE Global Communications Conference}, 2023, pp. 7055--7060.

\bibitem{Mota2021TheEO}
\BIBentryALTinterwordspacing
M.~P. Mota, A.~Valcarce, J.-M. Gorce, and J.~Hoydis, ``The emergence of wireless {MAC} protocols with multi-agent reinforcement learning,'' \emph{2021 IEEE Globecom Workshops (GC Wkshps)}, pp. 1--6, 2021. [Online]. Available: \url{https://api.semanticscholar.org/CorpusID:237091143}
\BIBentrySTDinterwordspacing

\bibitem{10817320}
B.~Madsen and R.~Adeogun, ``Federated multi-agent drl for radio resource management in industrial 6g in-x subnetworks,'' in \emph{2024 IEEE 35th International Symposium on Personal, Indoor and Mobile Radio Communications (PIMRC)}, 2024, pp. 1--7.

\end{thebibliography}
\end{document}